\documentclass[11pt]{article}
\usepackage{float}
\usepackage{cite}
\usepackage{amsmath,amssymb,amsfonts}
\usepackage{algorithmic}
\usepackage{algorithm}
\usepackage{graphicx}
\usepackage{textcomp}
\usepackage{booktabs}
\usepackage{epsfig,amsmath,amssymb,latexsym,color}
\usepackage{hyperref}
\usepackage{algorithm}
\usepackage{algorithmic}
\usepackage{multirow}
\usepackage{multicol}
\usepackage{wrapfig,lipsum,booktabs}
\usepackage{authblk}

\newcounter{mnote}
  \let\oldmarginpar\marginpar
 \renewcommand\marginpar[1]{\-\oldmarginpar[\raggedleft\footnotesize #1]%
    {\raggedright\footnotesize #1}}

\newtheorem{example}{Example}

\DeclareMathOperator*{\argmin}{arg\,min}

\begin{document}
\title{Local Clustering for Lung Cancer Image Classification via Sparse Solution Technique} 
\author{Jackson Hamel\footnote{jrhamel314@uga.edu. Department of Mathematics,
University of Georgia, Athens, GA 30602.} \quad\quad Ming-Jun Lai\footnote{mjlai@uga.edu. Department of Mathematics,
University of Georgia, Athens, GA 30602. This author is supported by the Simons Foundation 
Collaboration Grant \#864439.}\quad\quad  Zhaiming Shen\footnote{zhaiming.shen@uga.edu. Department of Mathematics,
University of Georgia, Athens, GA 30602.}\quad\quad Ye Tian\footnote{yt95681@uga.edu. Department of Mathematics,
University of Georgia, Athens, GA 30602.} 
}
\maketitle

\begin{abstract}
In this work, we propose to use a local clustering approach based on some sparse solution techniques to study the medical image classification problem, especially the lung cancer image classification task. 
We view images as the vertices in a weighted graph and compute the similarity between a pair of images as the edges in the graph. The vertices within the same cluster can be assumed to share similar features and properties, thus making the applications of graph clustering techniques very useful for image classification. Recently, the approach based on the sparse solutions of linear systems for graph clustering has been found to identify clusters more efficiently than traditional clustering methods such as spectral clustering. We propose to use the two newly developed local clustering methods based on sparse solution of linear system for image classification. However, a straight-forward application of the two methods does not work. In addition, we employ a box-spline-based tight-wavelet-frame method to clean up these images and help build a better adjacency matrix before clustering. These result an excellent method in classifying images. The performance of our method is significantly more efficient and either favorable or equally effective compared with other state-of-the-art approaches, e.g. convolutional neural network approach. Finally, we shall remark by pointing out two image deformation methods that can be used to build up more artificial image data to increase the number of labeled images.  
\end{abstract}


\section{Introduction}

Lung cancer is one of the most common and one of the deadliest cancers; only about $17\%$ of people in the U.S. diagnosed with lung cancer survive five years after the diagnosis. Currently, diagnostic methods to detect this cancer include biopsies and imaging, like CT scans. Given the severity of the cancer, early detection significantly improves the chances of survival, but it is also more difficult to detect early stages of lung cancer as there are typically fewer symptoms. To add to the difficulty and
complexity, when imaging techniques display cancerous-looking growths, they could either be benign (not harmful) or malignant (harmful). Thus, our task is a ternary classification problem to detect the presence of lung cancer in a patient's CT scans and determine whether a growth is benign or malignant. 

We are interested in providing a complementary method to help computer-aided clinic diagnosis (CACD) for lung cancer image classification. More precisely, we would like to 
help improve the computational feasibility, interpretability, and robustness of the existing methods in image-based CACD. 
The medical image analysis for lung cancer classification involves two types of approaches: traditional machine learning algorithms, which classify cancers based on manually extracted features from the images, and deep learning techniques, which automatically learn and extract features directly from the raw images based on tons of
labeled images. So far, the most popular methods to classify images are 
based on neural network structures such as ResNet50, EfficientNetB0, InceptionV3, MobileNetV2, DenseNet121, 
ResNet101, VGG18. See, e.g., \cite{Z21}, \cite{HM18}, \cite{Al21}, \cite{Raza23} for more details. In addition, topological machine learning algorithms such as \cite{A23} and \cite{Y23}
were recently developed to provide additional competitive and complementary methods.  
There are many other algorithms available in the literature and existing online.

The primary objective of this study is to classify lung cancer into three distinct categories: benign, malignant, and normal, based on CT scan slices of lung nodules without detecting cancerous cells from the CT scan images. To accomplish this task, we propose to use a simple yet effective local clustering approach based on the sparse solutions of linear systems for image classification. Our idea is based on the methods recently developed in 
\cite{LM20,LS23,LS23b}. See \cite{Shen24Diss} for a comprehensive study of these methods. 
The main idea of these methods is that we can view a set of image data as a graph
$G$ whose vertices are associated with images. Images in the same cluster are in the same class, e.g. benign,  malignant, or normal. Given a testing (query) image, we look for a cluster, usually of a small size, which contains the query image as the seed. For example, suppose the output cluster (excluding the query image) is of size 6. If all 6 images belong to one of the three classes, say malignant, the query image is concluded to be a malignant image.  Of course, we can also use a weaker criterion: the testing image is malignant if 3 out of the 6 images are malignant while 2 out of the 6 images are benign and 1 out of the 6 images are normal. One of the
significant differences of the graph clustering methods in \cite{LM20,LS23,LS23b} and the classic graph clustering method is that the former clustering methods are based on seeds and the size of the cluster containing the seeds. This significantly differs from a standard  size of clusters which are the size 
of the labeled images of benign, malignant, and normal cases.  

To perform the clustering task, the first and foremost step is to build up an effective adjacency matrix for the group of training images. There are a lot of distance functions available for building an adjacency matrix.  
As one needs to find the best one in the
sense that when the graph only based on labeled sample images, the adjacency matrix should have a block diagonal structure after the permutation according to the class membership.  Our method to generate a good adjacency matrix is introduced in Section \ref{secGenAdj} with more details. Mainly we will present a box spline based on tight-wavelet frames (TWF) to simplify all the images by removing some redundant information away so that the adjacency matrix built by a modified exponential distance (cf. \cite{ZP04}) is much cleaner. Although the TWF 
was constructed more than 10 years ago, they were mainly used for image compression/denoising/edge detection. Our study shows that they can be extremely useful for medical image classification.  

After building up the adjacency matrix, we simply form its graph Laplace matrix 
and use our local clustering techniques based on sparse solutions of linear systems (cf. \cite{LW21}). Let us 
explain our idea as follows.   Letting ${\bf c}_1$ be the indicator vector of cluster $C_1$ whose entries are 1 associated with the vertices in the cluster $C_1$, and 0 otherwise. It is easy to see that ${\bf c}_1$ is a sparse vector. For example, the cluster $C_1$ may have $\|{\bf c}_1\|_0 =5$ while the total number of 
images is possibly 
$N\gg 1000$. Thus, the number of nonzero entries in $C_1$, i.e. $\|{\bf c}_1\|_0$ is much smaller than the size $N$ of the graph.  Since we have 
\begin{equation}
{\cal L}{\bf c}_1=0,
\end{equation}
where ${\cal L}$ is the graph Laplacian of the graph $G$ (cf. \cite{C97}),  we cast the problem of finding ${\bf c}_1$ as a minimization problem
   \begin{equation}
   \label{sparsesol}
    \min\|{\bf x}\|_0,\ {\cal L}{\bf x}=0,\ {\bf x}\neq{\bf 0},  
    \end{equation}
where $\|{\bf x}\|_0$ stands for the number of nonzero entries in ${\bf x}$.  This problem was studied two decades ago (cf. \cite{Candes06} and \cite{Donoho06}) as a compressive sensing problem.  See also \cite{LW21} for a detailed explanation.  In this paper, we will present two computational algorithms specific to the local clustering problem for identifying lung cancer images. 
The details are left to Section \ref{seclocalClustering}.
 
We shall first demonstrate the effectiveness of these two computational algorithms in categorizing images of various human faces based on two well-known human face datasets: AT\&T and YaleB human face data. Then we will show how the proposed two algorithms work well for the medical image datasets we studied in this paper. 
One of the major advantages of our approach is its computational efficiency, as our approach only involved in finding a sparse solution of the linear system associated with the graph Laplacian matrix. It does not need to compute the eigenvectors of the graph Laplacian matrix associated with eigenvalue zero, which will be more costly when the size of the image data is large. 
Due to the given query image as a seed, we compute the only cluster that contains the seed.  
Our numerical 
experiment can be done in minutes while a convolutional neural network requires hours or more. We summarize our experimental results in Section \ref{secExp}.

The goal of our research is to determine a testing (query) image as one of the
three classes: Benign, Malignant, or Normal. Let us  summarize our computational procedure as follows: 
\begin{itemize}
\item Preprocess the query images by rescaling the images to the same size as the size of all labeled images and applying a Gaussian blur algorithm to reduce the high-frequency components of these images.
\item Simplify these images by using, e.g. PCA,  TWF (Tight Wavelet Frames) to be discussed in this paper, Watershed,  etc..
\item Build up an adjacency matrix $A$ by using, e.g. K-NN,  $L^2$ distance, modified exponential distance 
which is given in this paper.
\item For each testing image, we use either one of the two proposed local clustering methods: Local Cluster Extraction (LCE) or Least Squares Clustering (LSC), to find a 
small cluster of images similar to the given image.  
\item Make a conclusion about the label of the testing image based on the majority vote of the labels of the images in the found small cluster.  
\end{itemize}

We shall explain these steps of computation in the following sections.
{ The main contributions of this work are the following:}
\begin{itemize}
    \item We propose to use an approach based on sparse solution techniques for local clustering in order to identify a given query image to be benign, malignant, or normal.

    \item We treat each image as a vertex in the graph, and use modified exponential distance to build the adjacency matrix between all pairs of images. Then we apply the LCE or LSC for medical image classification (identification).

    \item The image is preprocessed by applying a box spline-based tight-wavelet frames (TWF) method, with edges being detected and enhanced.

    \item The experiments demonstrate that our image classification is fast and accurate, and hence can serve as a complementary method to help CACD. 
\end{itemize}

Furthermore, we notice that the size of the labeled data set we are using is small, about 1097 images.   To achieve a better result, we can increase the size of the labeled data. We propose two approaches: one is based on the solution of the optimal transport problem, i.e. a numerical approximation of the minimizer based on the solution of Monge-Amp\'ere equation studied in \cite{LL24} to help generate more labeled images.  The other approach is based on harmonic generalized barycentric coordinates (GBC) for image deformation (cf. \cite{DHL22} and \cite{H23}). 
We will present a few examples to illustrate the GBC approach for generating more labeled images at the end of the paper
while leaving the entire computation of generating new images to future work.  
Nevertheless, we will make some remarks in Section \ref{secRemark} to convince the readers 
that the proposed approach will generate enough labeled data.

\section{Image Simplification} \label{secImgSimp}
It is crucial to simplify the images so that we can keep the important features in an image while removing the noises before we feed them into any machine-learning model for image classification. It is well-known that we can use the standard PCA to simplify an image.  In fact, we have tried to use the first 10 singular values of each image to build up an adjacency matrix $A$, however, it 
turns out that the numerical results of image classification based on PCA simplification do not work very well. We thus turned to other approaches. 

\begin{figure}[!t]
\centering
\includegraphics[width=\columnwidth]{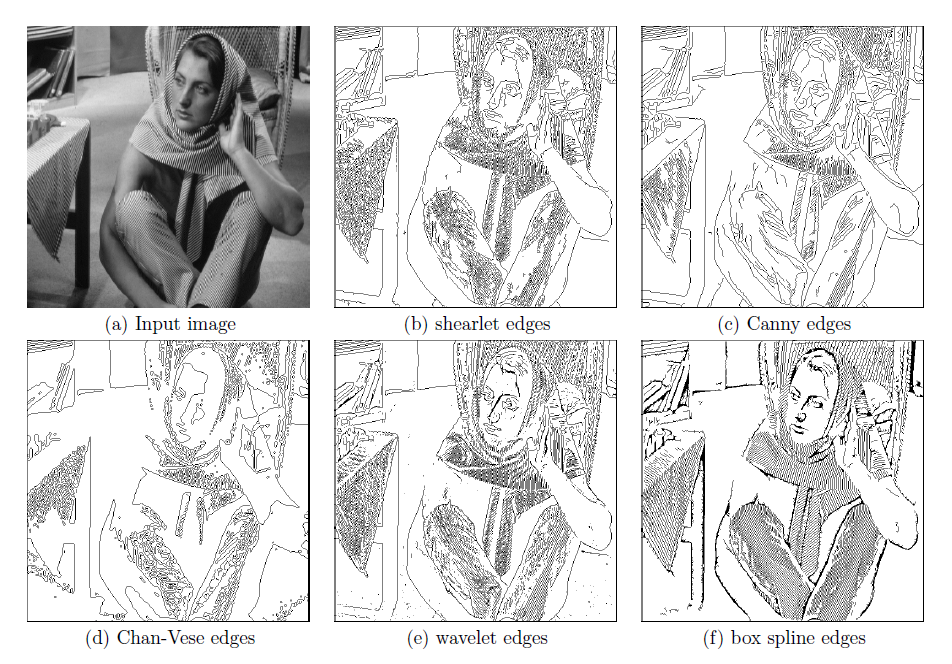}
\caption{An example of image simplification by using box spline-based tight-wavelet frames 
courtesy of Ming-Jun Lai, an author of \cite{GL13}. \label{edgesofbarbara}}
\end{figure}

We now explain the box spline-based tight-wavelet frames,  called LN method, 
to simplify these images.  The tight wavelet frames (TWF) we used  in this paper  were constructed in \cite{N05} based on integer translations of
box spline $B_{2211}$ (cf. \cite{BHR93} and \cite{LS07}).   Dr. Kyunglim Nam implemented the constructive method in \cite{LS06} based on box spline $B_{2211}$ to construct tight-wavelet framelets for image decomposition and reconstruction. In fact, she constructed tight-wavelet frames based on other box splines. See her dissertation \cite{N05} for more detail. Her MATLAB codes were further revised by Dr. Ming-Jun Lai with a denoising technique to have a package called LNmethod.m. As shown in Figure~\ref{edgesofbarbara}, we can easily see the box spline-based tight-wavelet frames method outperforms other traditional edge detection approaches and is able to capture and enhance the edge features across all directions in the image. More recent study can be found in \cite{GL13}. 

\begin{figure}[H]
	\centering
    \begin{tabular}{cc}
		\includegraphics[width=.45\columnwidth]{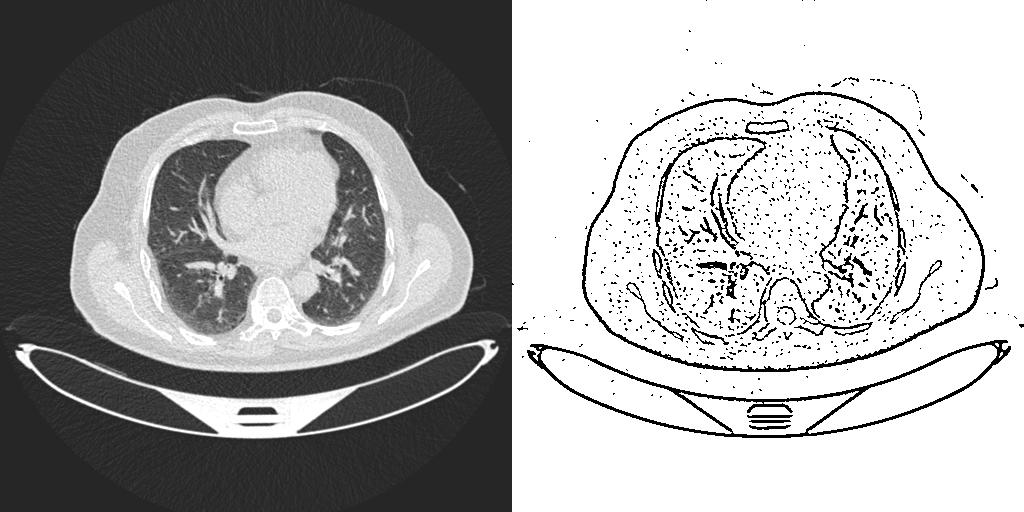} &		 	\includegraphics[width=.45\columnwidth]{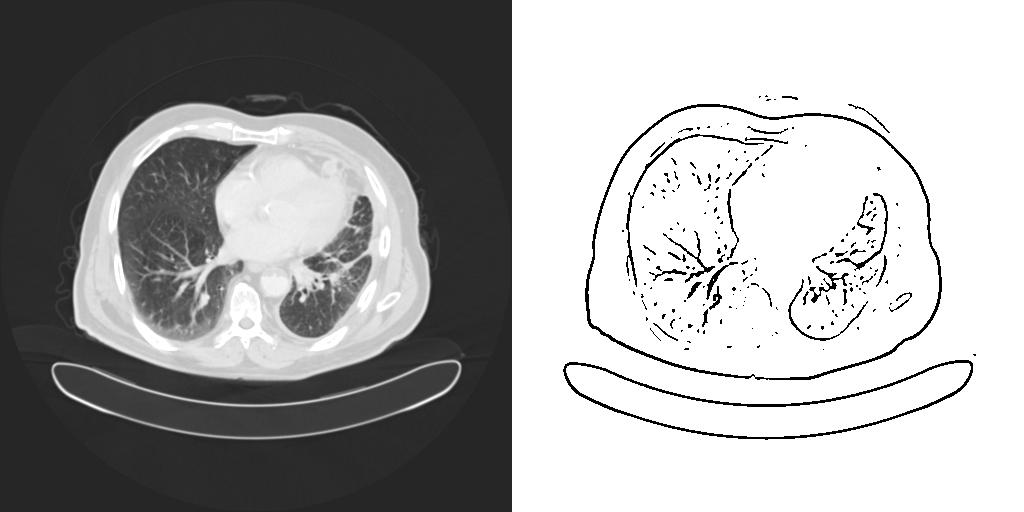}
    \end{tabular}
	\caption{Two examples of lung image simplification based on the LN method.} 
	 \label{LungAtwf}
\end{figure}

The main idea of the LN method is to decompose an image into its low-frequency part and high-frequency part by using the tight-wavelet frames and then reconstruct the high-frequency part back based only on the high-pass frequency part without the low-pass frequency. The resulting image is noisy and is denoised by using a cutoff parameter which is dependent on each image. One of our contributions in this paper is that we found an intelligent way to let the computer decide the best parameter for cleaning each of the medical images of interest. Instead of manually choosing a cut-off parameter for each image, we let computer decide the cutoff by computing the percentage of the density of an image before and after the simplification and make sure that the percentage is within the given range. This way we algorithmically clean up more than 1000 images in a few minutes. Let us present some examples of medical image simplification by using the LN method in Figure~\ref{LungAtwf}. 


\section{Generation of Adjacency Matrices} \label{secGenAdj}
Let us now explain the way we build the adjacency matrices for our graph clustering approach. We mainly use 
the distance associated with the Gaussian kernel with different parameters to generate a working adjacency matrix.
Let us call it \emph{Modified Exponential Distance:} 

Let ${\bf x}_i \in \mathbb{R}^n$ be the vectorization of each image from the original data set
and $ NN({\bf x}_i, K)$ is the set of K-nearest neighbors of ${\bf x}_i$. For $r\ge 1$, let 
$\sigma_i= \|{\bf x}_i- {\bf x}^{i,r}\|$ be the $r$-th closest point of ${\bf x}_i$.  We build the adjacency matrix associated with the data set to be $A=[A_{ij}]_{i,j=1,\cdots, n}$ with entries
\begin{equation} \label{MED}
    A_{ij}= \left\{ \begin{array}{cc} \exp(- \|{\bf x}_i - {\bf x}_j\|/(\sigma_i \sigma_j) & \hbox{ if } {\bf x}_j \in 
NN({\bf x}_i, K)\cr 0 & \hbox{otherwise } \end{array} \right. 
\end{equation}
for all $i,j=1, \cdots, n$  (cf. \cite{ZP04}).

Note that the above $A_{ij}$ is not necessarily symmetric, so
we consider $\tilde{A}_{ij} = A^\top A$ for symmetrization. 
Alternatively, one may also use 
$\tilde{A}_{ij} = \max\{A_{ij}, A_{ji}\}$ or $\tilde{A}_{ij} =(A_{ij} + A_{ji})/2$.  


Based on the modified exponential distance (\ref{MED}), we generate the adjacency matrix with the choice of two parameters $r$ and $K$ over all the images.  We visually examine the adjacency matrices for various $r=2, 
\cdots, 10$ and $K=5,\cdots, 15$ and identify 
the best pair $(r,K)$ which has a block diagonal structure, 
where each block corresponds to one class in the dataset.
\begin{figure}[!t]	
	\centering
	
 \begin{tabular}{cc}
 \includegraphics[width=0.45\textwidth]{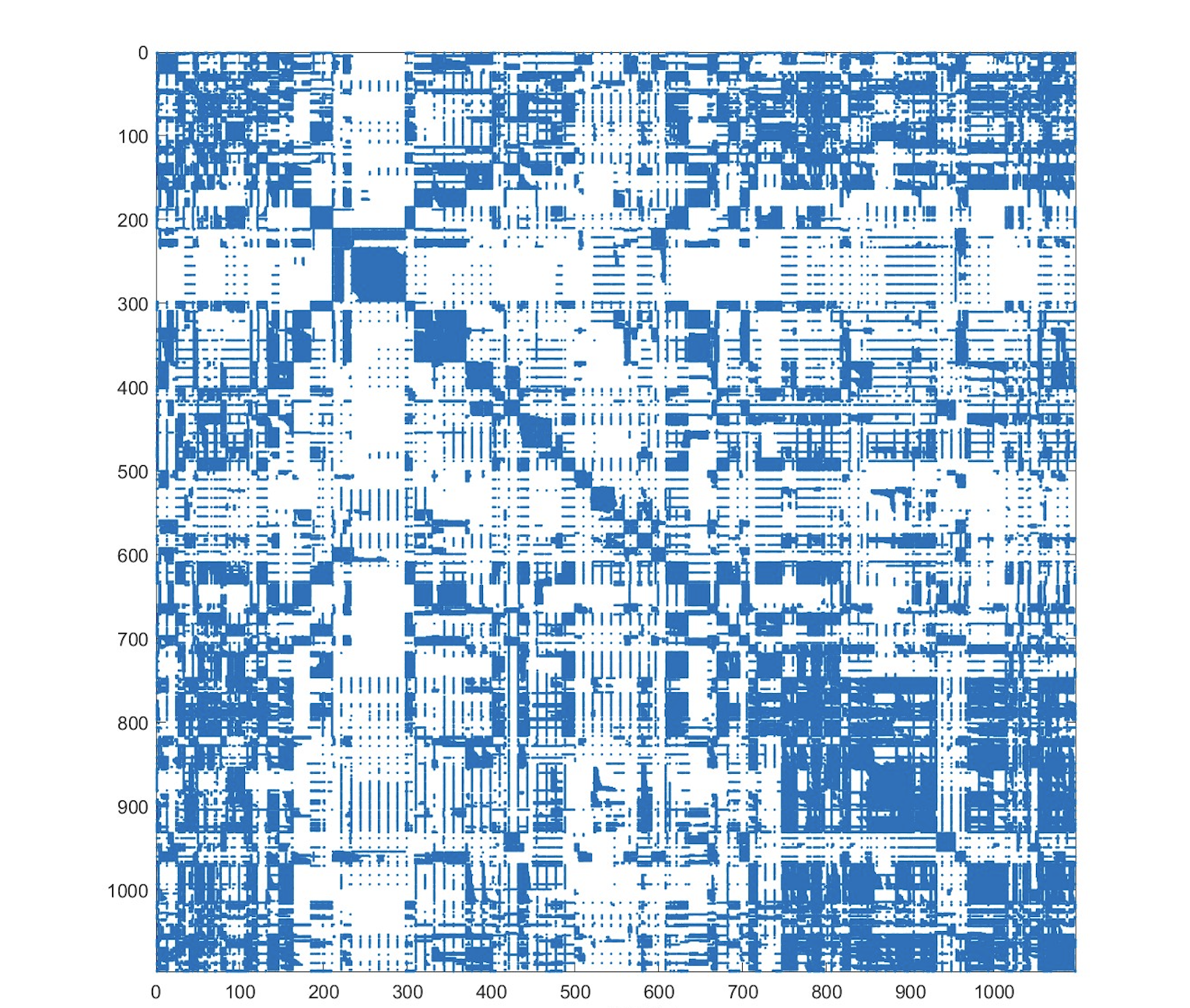}  &     \includegraphics[width=0.45\textwidth]{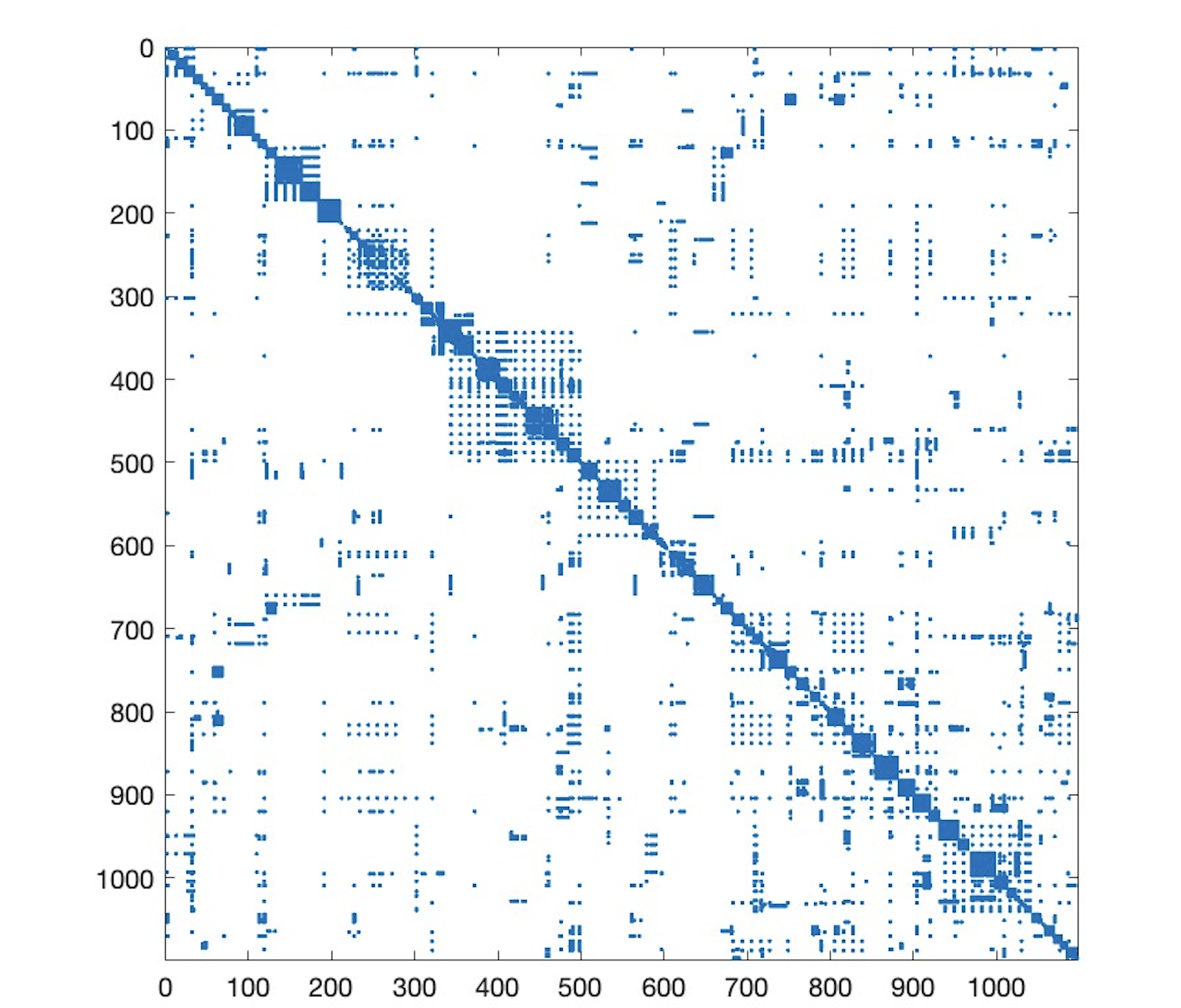}
 \end{tabular}	
	\caption{Adjacency matrices built without (left) and with the LN method (right) based on the Modified Exponential Distance.} 
	 \label{LungA}
\end{figure}
The two adjacency matrices, without and with the LN method, are shown in Figure~\ref{LungA}.  One can easily see that the adjacency matrix based on the LN method is much cleaner than the adjacency matrix built from the original images.


\section{Local Clustering Based on Sparse Solution Technique} \label{seclocalClustering}

Let us explain the idea of two local clustering methods  \cite{LS23} and \cite{LS23b} in this section. Recall that if a graph $G$ has $k$ connected components $C_1,\dots,C_k$, then its graph Laplacian $\mathcal{L}$ can be written into block diagonal form
 
\begin{equation}
    \mathcal{L} = 
    \begin{bmatrix}
        \mathcal{L}_1 & & &  \\
        & \mathcal{L}_2 & &  \\
        & & \ddots & \\
        & & & \mathcal{L}_k
    \end{bmatrix},
\end{equation}
\noindent 
with each $\mathcal{L}_i$ being the graph Laplacian of the $i$-th component.  It is known that the indicators ${\bf c}_1, \cdots, {\bf c}_k$, which are associated with the clusters $C_1, \cdots, C_k$,
are in the null space of the kernel of $\mathcal{L}$.  That is, $\mathcal{L} {\bf c}_i=0, i=1, \cdots, k$ (cf. \cite{C97} and \cite{L07}). 

Note that $\|{\bf c}_i\|_0 \ll n$,   where $\|{\bf x}\|_0$ is the number of nonzero components in ${\bf x}$. Then we can find such component by solving the following minimization  problem:
\begin{equation} \label{L0min}
    \min\|{\bf x}\|_0,\ {\cal L}{\bf x}=0,\ {\bf x}\neq{\bf 0}.
\end{equation}
\noindent
In practice, the graphs are usually "noised", and thus we may not be able to find an exact solution ${\bf x}$. However, as long as the noise is small enough, the output solution based on the perturbed version of $\mathcal{L}$ should be close to the exact solution.  


To avoid getting zero vector as a solution, we also move a collection $T$ of columns associated with seeds, in fact one
column associated with the query image in this paper, and solve for 
    \begin{equation}
    \label{L2min}
    \argmin_{{\bf x}\in\mathbb{R}^{|V|-|T|}}\{\|\mathcal{L}_{V\backslash T}{\bf x}-{\bf y}\|_2: \ \|{\bf x}\|_0\le s\}, 
    \end{equation}
where $s$ is the sparsity constraint assumption for ${\bf x}$, and   ${\bf y}$ is the row sum vector of $\mathcal{L}_{V\backslash T}$.
We present the detailed procedure for this method, named Local Cluster Extraction (LCE), as Algorithm \ref{algCSLCE}.

\begin{algorithm}[h]
    \caption{Local Cluster Extraction \cite{LS23b}}
    \label{algCSLCE}
    \textbf{Input}: Adjacency matrix $A$, and a small set of seeds $\Gamma\subset C_1$ \\
    \textbf{Parameter}: Estimated size $\hat{n}_1\approx |C_1|$, random walk threshold parameter $\epsilon\in (0,1)$, random walk depth $t\in\mathbb{Z}^{+}$, sparsity parameter $\gamma\in [0.1, 0.5]$, rejection parameter $R\in [0.1, 0.9]$.\\
    \textbf{Output}: The target cluster $C_1$
    \begin{algorithmic}[1] 
        \STATE Compute $P=AD^{-1}$,  $\mathbf{v}^{0}=D\mathbf{1}_{\Gamma}$, and $L=I-D^{-1}A$.
        \STATE Compute $\mathbf{v}^{(t)}=P^t\mathbf{v}^{(0)}$.
        \STATE Define $\Omega={\mathcal{L}}_{(1+\epsilon)\hat{n}_1}(\mathbf{v}^{(t)})$. 
        \STATE Let $T$ be the set of column indices of $\gamma\cdot|\Omega|$ smallest components of the vector $|L_{\Omega}^{\top}|\cdot|L\mathbf{1}_{\Omega}|$.
        \STATE Set $\mathbf{y}:=L\mathbf{1}_{V\setminus T}$. Let $\mathbf{x}^\#$ be the solution to
        \begin{equation} 
        \argmin_{\mathbf{x}  \in \mathbb{R}^{|V|-|T|}} \{\|L_{V\setminus T}\mathbf{x}- \mathbf{y}\|_2: \|\mathbf{x}\|_0\leq (1-\gamma)\hat{n}_1\}
        \end{equation}
        obtained by using $O(\log n)$ iterations of \emph{Subspace Pursuit} \cite{DM09}.
        \STATE Let $W^{\#} = \{i: \mathbf{x}_i^{\#}>R\}$  . 
        \STATE \textbf{return} $C^{\#}_1=W^{\#}\cup T$. 
    \end{algorithmic}
\end{algorithm}



Another similar computational algorithm proposed in \cite{LS23} is to simply drop the sparsity constraint in (\ref{L2min}), hence we will solve a least squares problem. We name the method Least Squares Clustering (LSC) and present it as Algorithm \ref{alg:cluster_pursuit}. There are many other algorithms available that solve the minimizations (\ref{L0min}) and (\ref{L2min}). We refer the interested reader to several other methods in \cite{LW21}. 

\begin{algorithm}[h]
    \caption{Least Squares Clustering \cite{LS23}}
    \label{alg:cluster_pursuit}
    \textbf{Input}: Adjacency matrix $A$, and a small set of seeds $\Gamma\subset C_1$. \\
    \textbf{Parameter}: Estimated size $\hat{n}_1\approx |C_1|$, random walk threshold parameter $\epsilon\in (0,1)$, random walk depth $t\in\mathbb{Z}^{+}$, least squares threshold parameter $\gamma\in (0,1)$, rejection parameter $R\in [0.1, 0.9]$.\\
    \textbf{Output}: The target cluster $C_1$
    \begin{algorithmic}[1] 
    \STATE Compute $P=AD^{-1}$,  $\mathbf{v}^{0}=D\mathbf{1}_{\Gamma}$, and $L=I-D^{-1}A$.
        \STATE Compute $\mathbf{v}^{(t)}=P^t\mathbf{v}^{(0)}$.
        \STATE Define $\Omega={\mathcal{L}}_{(1+\epsilon)\hat{n}_1}(\mathbf{v}^{(t)})$. 
        \STATE Let $T$ be the set of column indices of $\gamma\cdot|\Omega|$ smallest components of the vector $|L_{\Omega}^{\top}|\cdot|L\mathbf{1}_{\Omega}|$.
        \STATE Set $\mathbf{y}:=L\mathbf{1}_{\Omega\setminus T}$. Let $\mathbf{x}^\#$ be the solution to
\begin{equation} 
\argmin_{\mathbf{x}  \in \mathbb{R}^{|\Omega|-|T|}} \|L_{\Omega\setminus T}\mathbf{x}- \mathbf{y}\|_2
\end{equation}
obtained by using an iterative least squares solver. 
        \STATE Let $W^{\#} = \{i: \mathbf{x}_i^{\#}>R\}$.
        \STATE \textbf{return} $C^{\#}_1=\Omega\setminus W^{\#}$.
    \end{algorithmic}
\end{algorithm}



Let us first present two examples of clustering human faces to demonstrate that our local clustering based on sparse solution methods works very well before we present its performance on the medical images. 
One of the main contributions in this paper is that we design the procedure to recover small clusters instead of the cluster for the entire class. Both LCE and LSC perform better when sparsity of the indicator vector, i.e., the size of the cluster, is small relative to the size of the whole dataset. Moreover, it is difficult for the resulting auxiliary matrix to perfectly recover adjacency information among all images of the same class. In our experiment, we found that the LN method together with modified exponential distances performs well in identifying smaller clusters belonging to the same classes, as shown in Fig.\ref{LungA}, where small blocks along the diagonal indicate clusters. Therefore, in our algorithm \ref{algII}, we seek to identify clusters of small sizes.

\begin{figure}[t]	
	\centering
	\begin{tabular}{cc}	
 \includegraphics[width=0.45\columnwidth]{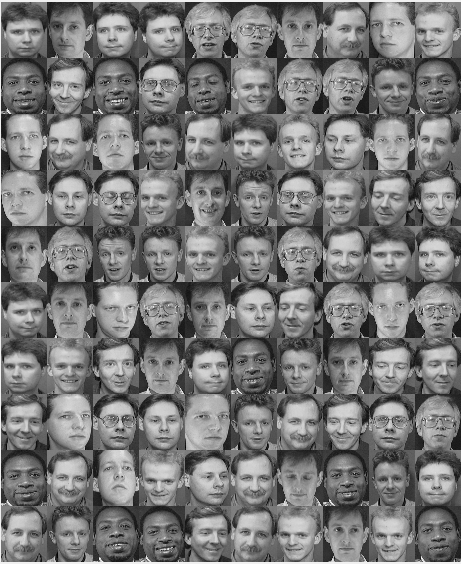} &
		\includegraphics[width=0.45\columnwidth]{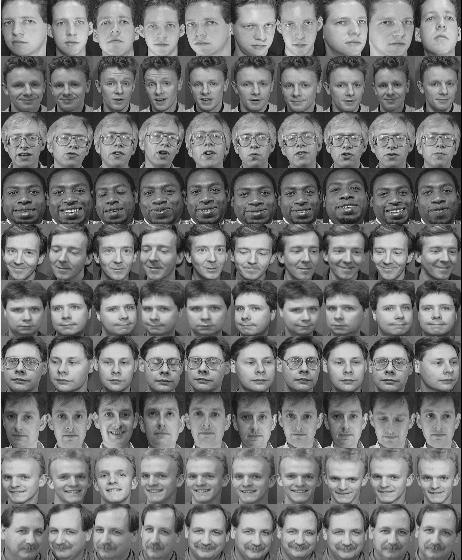}
	\end{tabular}
  \vspace{-1mm}
	\caption{Given a set of faces (left), we sorted these faces (right) which were done by using LCE or LSC. \label{facesorting}}
\end{figure}

\begin{example}
    The AT\&T human faces dataset \cite{SH94} contains grayscale images for $40$ different people of pixel size $56\times 46$. Images of each person are taken under $10$ different conditions, by varying the three perspectives of faces, lighting conditions, and facial expressions. As shown in Figure~\ref{facesorting}, we use part of this dataset and  the modified exponential distance function explained before and then apply the LCE and LSC methods to extract the faces into their correct clusters.
\end{example}
 
\begin{figure}[htpb]	
 	\centering
	\begin{tabular}{cc}	
 		\includegraphics[width=0.45\columnwidth]{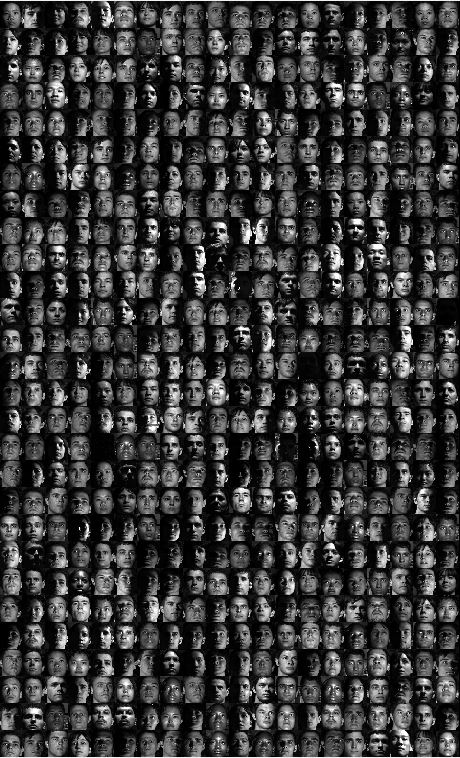} & \includegraphics[width=0.45\columnwidth]{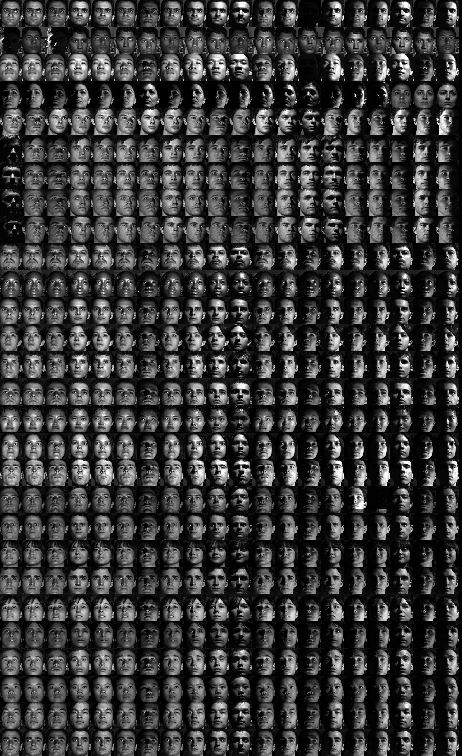}
	\end{tabular}  
 	\caption{Given a set of faces (left), we sorted these faces (right) which were done by using LCE or LSC.} \label{YaleBfacesorting}
 \end{figure}

\begin{example}
    The "Extended Yale Face Database B (YaleB)" \cite{GBK01} dataset contains 16128 grayscale images of 28 human subjects under 9 poses and 64 illuminations. We use the modified exponential distance function explained before and then apply the LCE and LSC methods to extract the faces into their correct clusters. See Figure~\ref{YaleBfacesorting} for the illustration.
\end{example}

     

\section{Experiments}  \label{secExp}

\subsection{Dataset Description}
\begin{figure*}[h]
        \includegraphics[width=0.16\columnwidth]{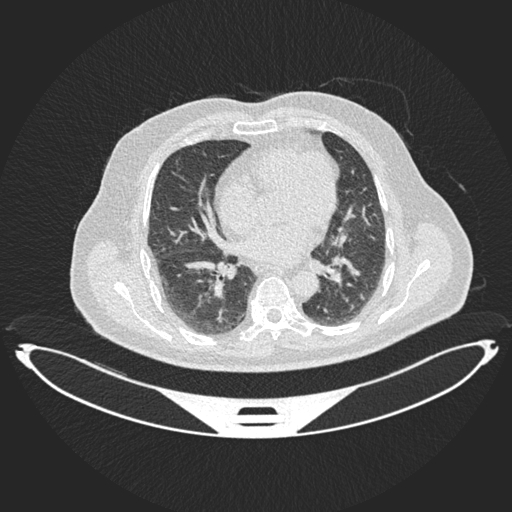} \includegraphics[width=0.16\columnwidth]{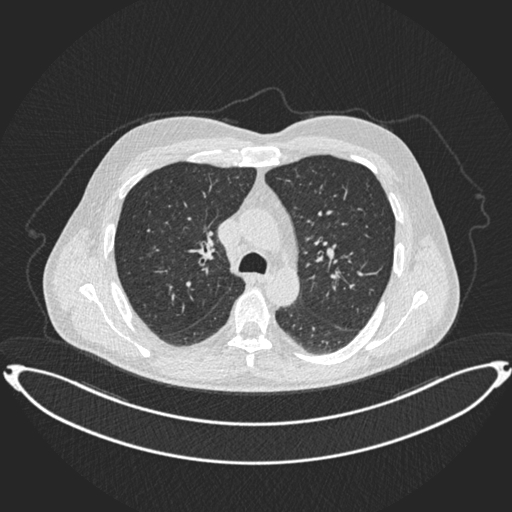}
        \includegraphics[width=0.16\columnwidth]{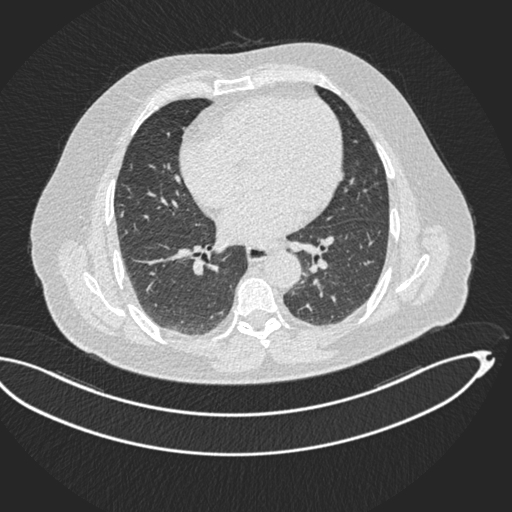} \includegraphics[width=0.16\columnwidth]{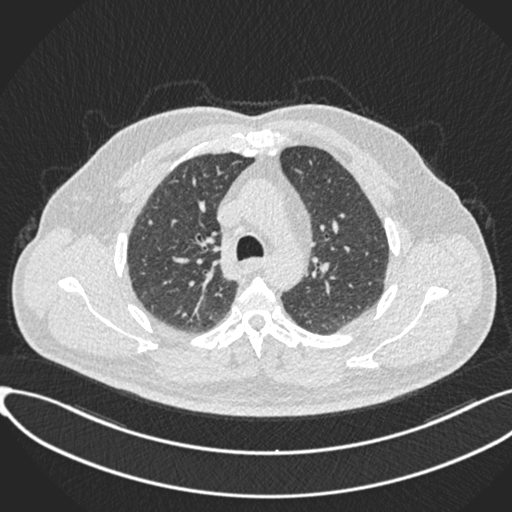} \includegraphics[width=0.16\columnwidth]{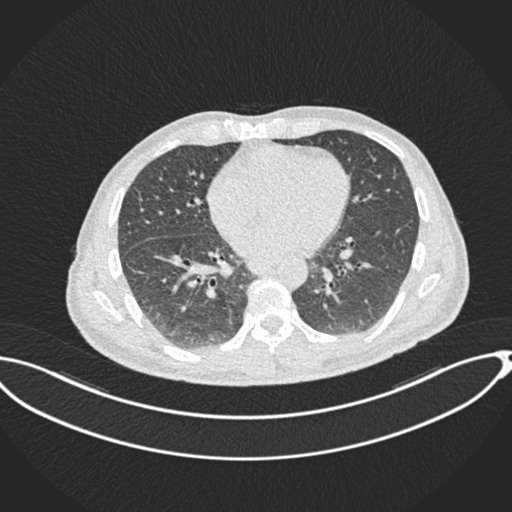} \includegraphics[width=0.16\columnwidth]{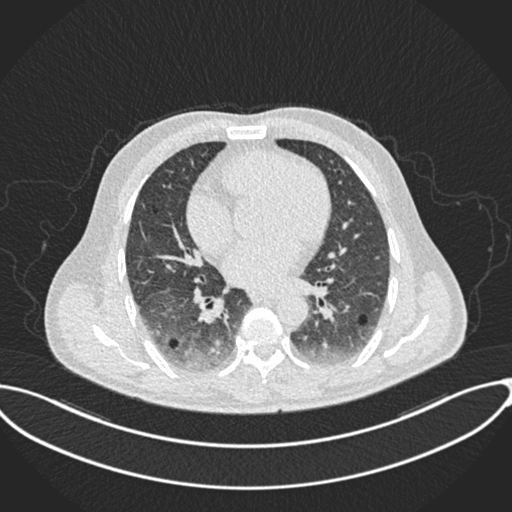}  \\
        \includegraphics[width=0.16\columnwidth]{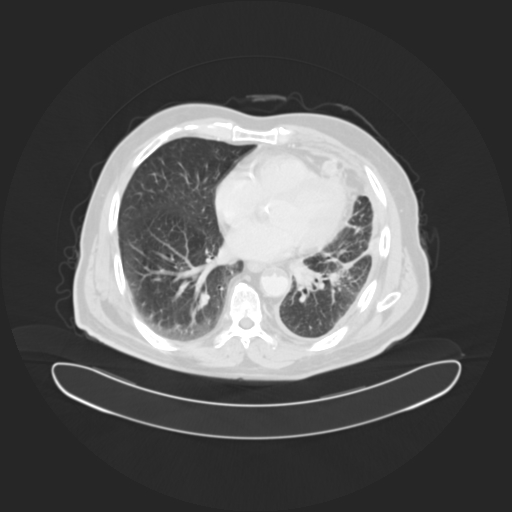} 
        \includegraphics[width=0.16\columnwidth]{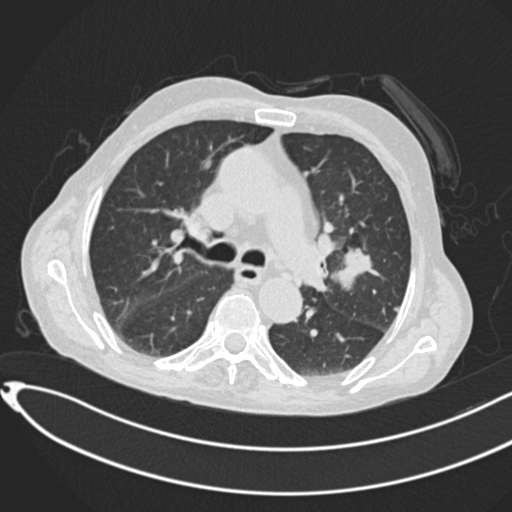} 
        \includegraphics[width=0.16\columnwidth]{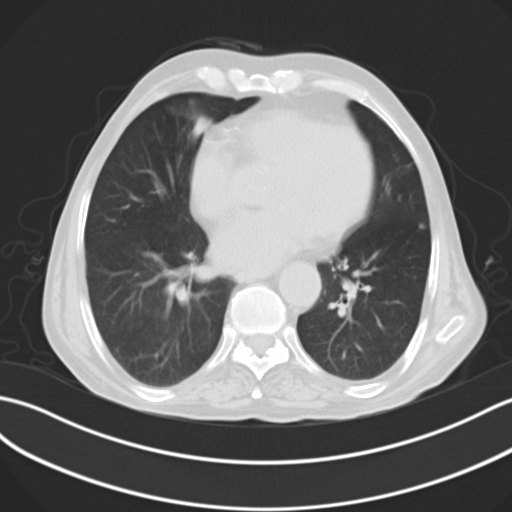} 
        \includegraphics[width=0.16\columnwidth]{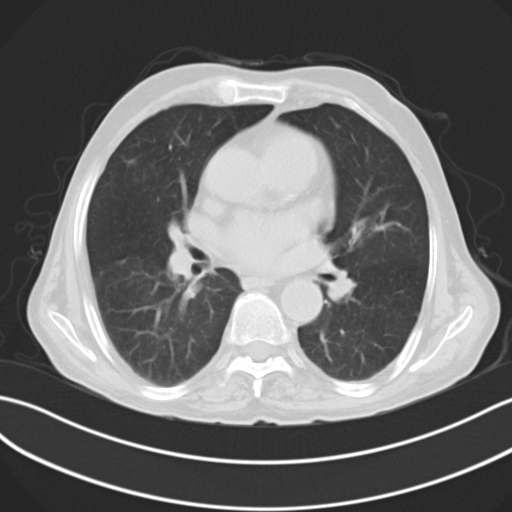} 
        \includegraphics[width=0.16\columnwidth]{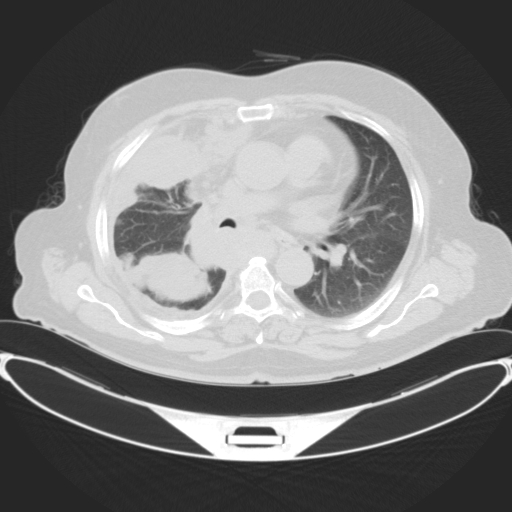}
        \includegraphics[width=0.16\columnwidth]{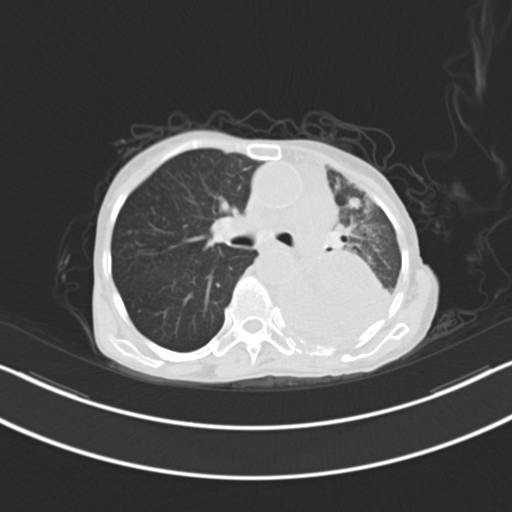} 
        \\
        \includegraphics[width=0.16\columnwidth]{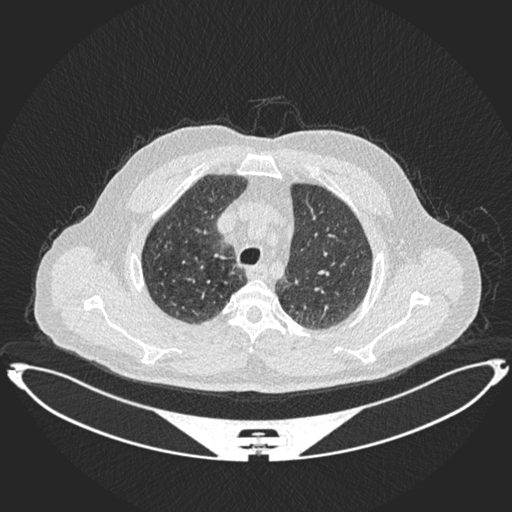} 
        \includegraphics[width=0.16\columnwidth]{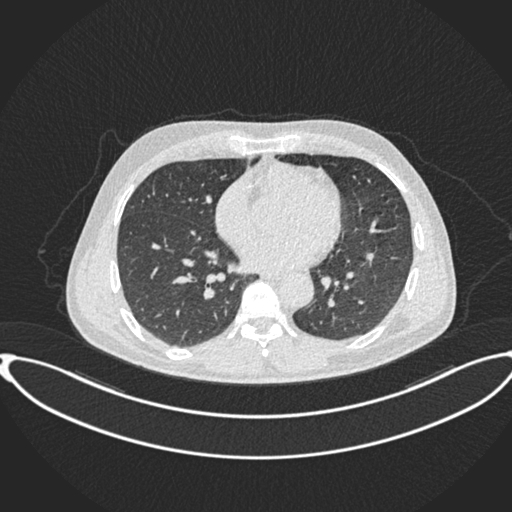} 
        \includegraphics[width=0.16\columnwidth]{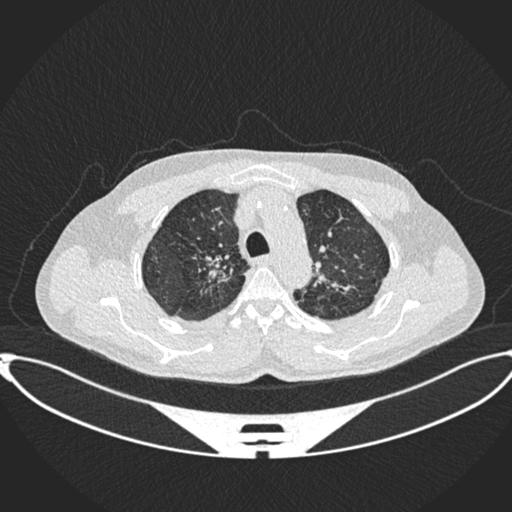} 
        \includegraphics[width=0.16\columnwidth]{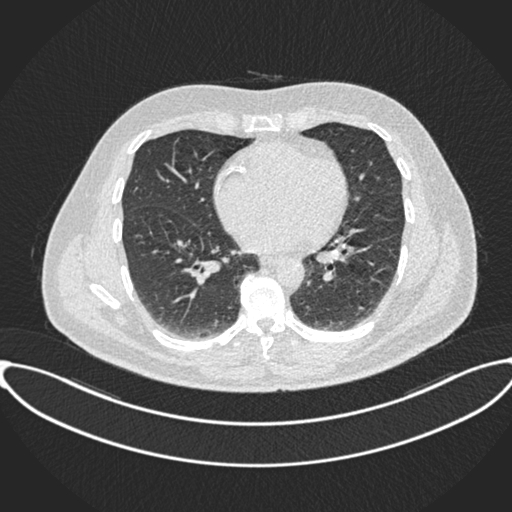} 
        \includegraphics[width=0.16\columnwidth]{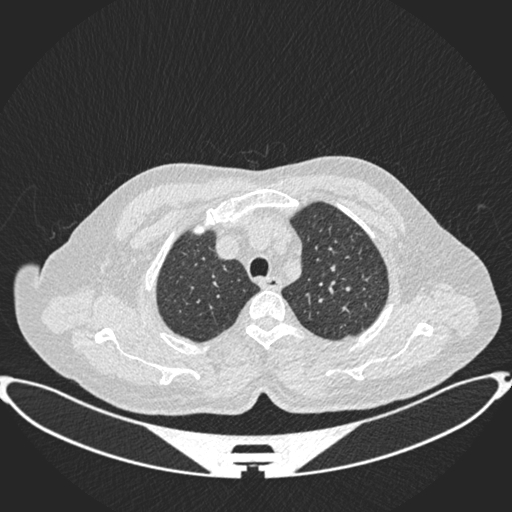}
        \includegraphics[width=0.16\columnwidth]{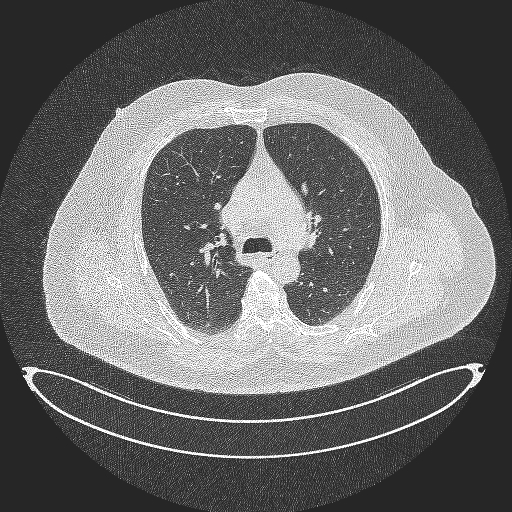}

        \caption{ Examples of benign cases (top row), malignant cases (middle row), and normal cases (bottom row). \label{imageexamples}}
    \end{figure*}

We carried out the experiments on the ``Iraq-Oncology Teaching Hospital/National Center for Cancer Diseases (IQ-OTH/NCCD)" lung cancer dataset. The dataset was collected from the “Iraq-Oncology Teaching Hospital and the National Center for Cancer Diseases” for more than three months in 2019. It consists of CT scans from patients, including both healthy individuals and those diagnosed with various stages of lung cancer. The dataset, annotated by multiple oncologists and radiologists, consists of 1,097 CT scan images of the human chest. These images represent 110 cases (40 identified as malignant, 15 as benign, and 55 as normal), with variations in age, gender, educational background, residence, and living conditions. The cases in this study were divided into three categories: benign, malignant, and normal, as depicted in Figure~\ref{imageexamples}. The IQ-OTH/NCCD dataset can be downloaded from the website such as \cite{Al20} on Kaggle. The breakdown of the distribution of this dataset by class-wise category is presented in Table~\ref{tab0}.

\begin{table}[t]
\centering
\caption{The class-wise distribution of the ``IQ-OTH/NCCD" lung cancer dataset. \label{tab0}}

\begin{tabular}{lcc}
\toprule
          Class  & Patients & No. of Samples \\
    \midrule
    Benign & 15 & 120 \\
    Malignant & 40 & 561 \\
    Normal & 55  & 416 \\
    Total & 110 & 1097 \\
	\bottomrule

\end{tabular}
\end{table} 


\begin{figure*}[h]	
	\centering
	\begin{tabular}{cc}	
 \includegraphics[width=0.45\columnwidth]{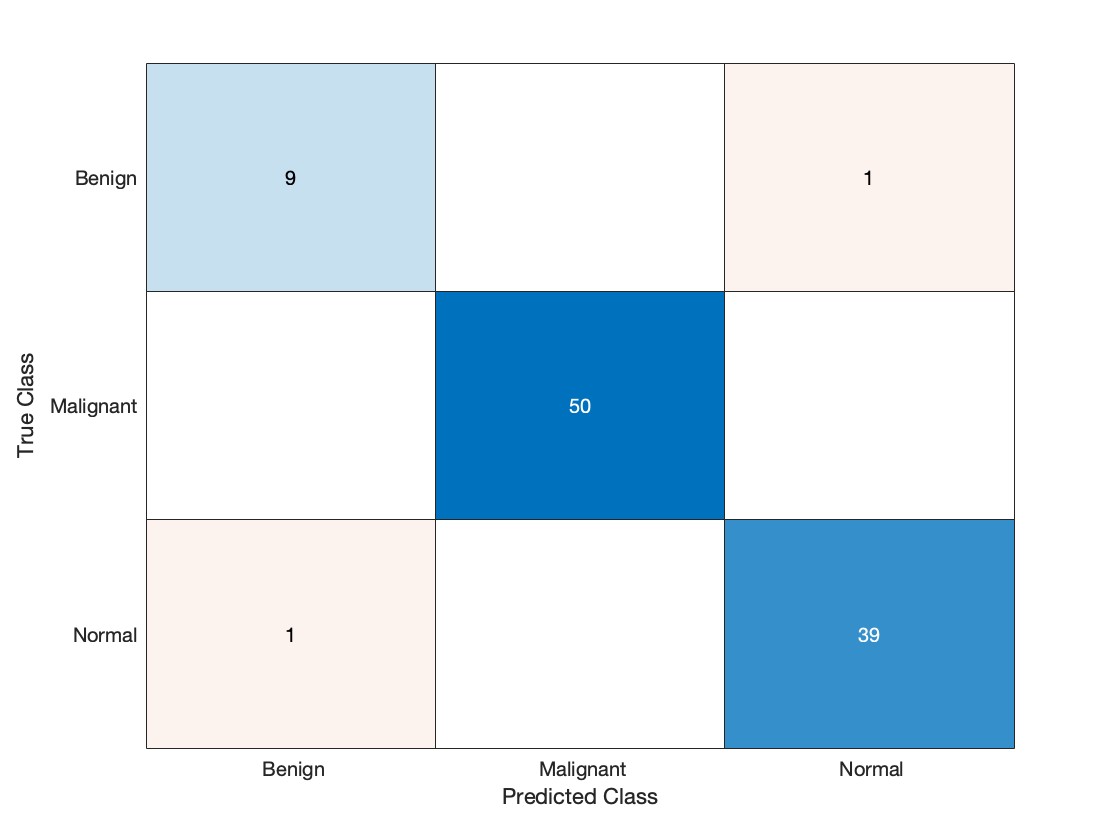} &
		\includegraphics[width=0.45\columnwidth]{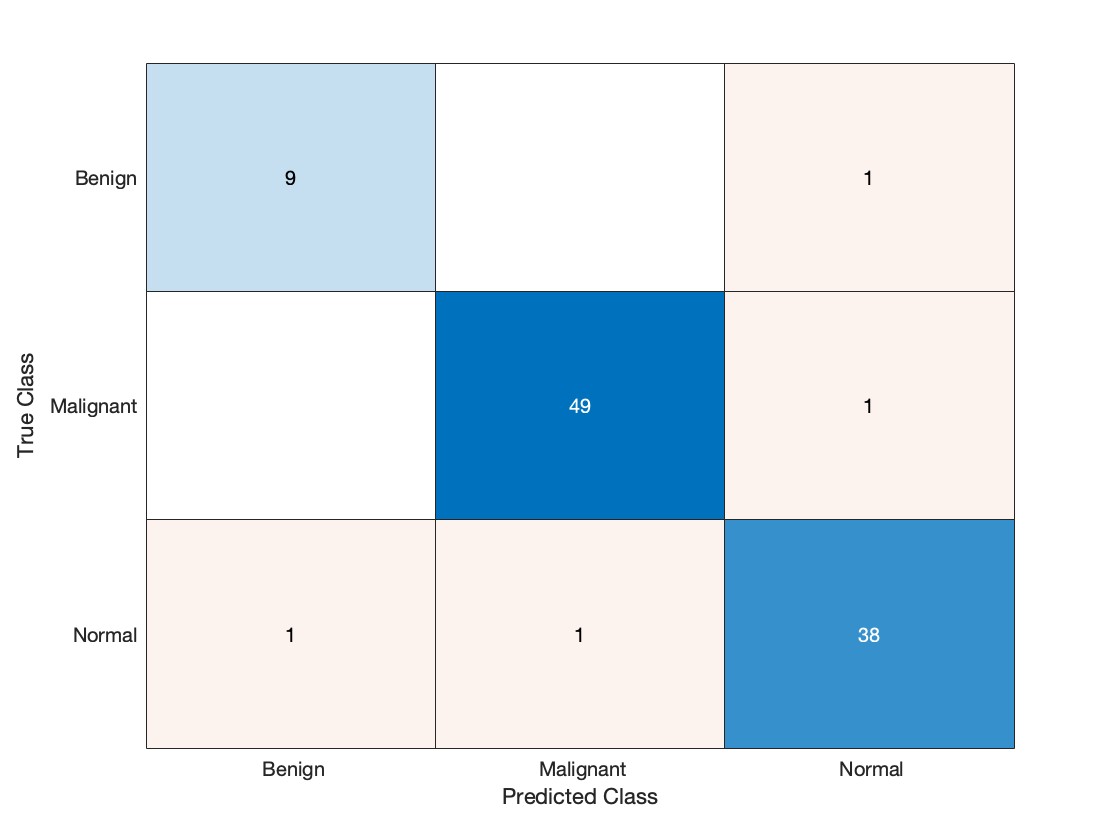}
	\end{tabular}
  \vspace{-3mm}
	\caption{Confusion matrix of LCE (left) and LSC (right) on IQ-OTH/NCCD dataset. The values are rounded to integers. \label{Cmatrix}}
\end{figure*}

\subsection{Experimental Procedure}
Let us now explain the experimental procedure to identify whether a given lung cancer image is benign, malignant, or normal.

First of all, we preprocess the data by scaling all the images into the standard size of $512\times 512$ as the given image data are not in a uniform size. Then we experimented with various image simplification methods, e.g. PCA, LN method, Watershed, etc.. It turns out that the
LN method can work nicely.  One of the mathematical heuristics is that each image has a lot of redundant information, e.g. the pixel values outside of the cross-section of the lung and tumor region which may inference the similarity of the two images.  Removal of these information is done by the LN method. 

After preprocessing and simplifying the image data, we then build the adjacency matrix using modified exponential distance. Let us outline our algorithm for identification of each individual image. Our image identification procedure is summarized as Algorithm \ref{algII}. 

\begin{algorithm}[h]
\caption{Image Identification}
    \label{algII}

    \begin{algorithmic}[1] 
        \STATE Given a testing (query) image S, we build the adjacency matrix by using modified exponential distance between the testing image and the remaining images with known labels.
\STATE Apply the LCE or LSC algorithm with the seed S to output a cluster $C_S$ of small size, e.g., size of 10.  
\STATE Conclude that $S$ belong to a certain class (benign, malignant, normal) if that class label appears the most (after excluding $S$ itself) in $C_S$.
    \end{algorithmic}
\end{algorithm}







We apply LCE or LSC on the adjacency matrix which is built by using the modified exponential distance between each testing image and the remaining images with known labels.
In the experiments, we randomly select 10 benign images, 50 malignant images, 40 normal images (roughly follow the distribution of the dataset), a total of 100 images out of 1097 images as testing images. For each individual testing image, we perform Algorithm~\ref{algII} to determine its class label. Note that the last step in Algorithm~\ref{algII} is based on the majority vote. For example, if the cluster (after excluding the testing image itself) is of size 10, which contains 2 images with label benign, 5 images with label malignant, 3 images with label normal, then this testing image is classified as malignant. The experiments are conducted over 100 repetitions. The code which can reproduce our experimental results is available on
\url{https://github.com/zzzzms/LocalClustering4LungCancer}. 


\subsection{Evaluation Metrics and Benchmarks}
We propose to use evaluation metrics, including Accuracy, Precision, Recall, F1-score, and confusion matrix to validate the performance of our approach. Comparisons are conducted on benchmark methods including the 
KNN (k-Nearest neighbors) \cite{Ab20}, integrated contrast/features-based method (CFM) \cite{Khan20}, weakly supervised deep learning (WSDL) \cite{Wang19}, Co-learning feature fusion maps (CLFM) \cite{Kumar19}, DFD-Net \cite{Sori21}, and CNN-SOA \cite{Yan23}.

\subsection{Hyperparameters}
For both the IQ-OTH/NCCD and Pneumonia dataset, we set $K=5$ and $r=3$ to generate the adjacency matrix based on modified exponential distance.

For LCE method on IQ-OTH/NCCD dataset, we set random walk threshold parameter to be 0.8, random walk depth to be 3, sparsity parameter to be 0.2, rejection parameter to be 0.0. The estimated size of output cluster is set to be 6. For LCE method on Pneumonia dataset, everything else is the same except that the output cluster is set to be 8.

For LSC method on both IQ-OTH/NCCD and Pneumonia dataset, we set random walk threshold parameter to be 0.2, random walk depth to be 3, sparsity parameter to be 0.2, least squares threshold parameter to be 0.2, rejection parameter to be 0.9. The estimated size of output cluster is set to be 10. 

\subsection{Results}
We show the average confusion matrix over 100 runs of LCE and LSC methods in Figure~\ref{Cmatrix}, note that the entries in the confusion matrix are rounded to be integers.
The overall accuracy ($\pm$ standard deviation) for LCE and LSC on Pneumonia dataset are $97.10\%\pm 1.71\%$ and $95.56\%\pm 2.10\%$ respectively. The performances of the two proposed methods are presented in Table~\ref{tab1}, and the comparisons with the other benchmarks are summarized in Table~\ref{tab2} (best results are in bold and second best results are underlined). Note that the precision, recall, and F1-score in Table~\ref{tab2} are the macro-average values over the entire dataset.

    



    



\begin{table}[t]
\centering
\caption{Precision/Recall/F1-score of Image Identification by Two Local Clustering Methods  \label{tab1}}
\begin{tabular}{lcc}
\toprule
             & LCE & LSC  \\
    \midrule
    Benign & 87.98 / 87.80 / 87.89 & 85.38 / 88.80 / 87.06 \\
    Malignant & 98.90 / 99.34 / 99.12 & 97.59 / 98.10 / 97.84 \\
    Normal & 96.86 / 96.37 / 96.61 &  95.65 / 94.08 / 94.86 \\

	\bottomrule

\end{tabular}
\end{table} 

\begin{table}[t]
\centering
\caption{Performance of the two local clustering methods compared with other benchmarks  \label{tab2}}
\begin{tabular}{lcccc}
\toprule
            & Accuracy & Precision & Recall & F1-score \\
    \midrule
    
	LCE (ours) & \textbf{97.10} & \textbf{94.58} & \underline{94.50} & \textbf{94.54} \\
	LSC (ours) & 95.56 & \underline{92.87} & 93.66 & \underline{93.25} \\
        KNN \cite{Ab20} & 85.25 & 71.09 & 85.32 & 82.56 \\
        CFM \cite{Khan20} & 87.24 & 75.94 & 87.24 & 84.19 \\
        WSDL \cite{Wang19} & 89.26 & 78.26 & 89.16 & 86.37 \\
        CLFM \cite{Kumar19} & 92.46 & 81.39 & 91.08 & 88.34 \\
        DFD-Net \cite{Sori21} & 94.52 & 83.50 & 93.65 & 89.90 \\
        CNN-SOA \cite{Yan23} & \underline{96.58} & 84.16 & \textbf{95.38} & 91.35 \\

	\bottomrule

\end{tabular}
\end{table}


Furthermore, we compare our methods with several other modern deep learning approaches such as EfficientNetB0, GoogleLeNet, ResNet50, EfficientNetB4,  Attention-InceptionResNet-V2, and MobileNetV2. According to \cite{Raza23}, the testing accuracy for these deep neural network based approaches are $93.67\%$, $94.38\%$, $94.40\%$, $97.29\%$ $97.41\%$, $98.28\%$ respectively. The 
accuracy of LCE and LSC in Table~\ref{tab2} are comparable to these state-of-the-art results.  It is worthwhile to point out that the proposed methods LCE and LSC require no training and the identification of 100 testing images only takes about a few seconds, while the neural network approaches usually take about hours for training plus testing.

\section{Final Remarks} \label{secRemark}
Let us conclude the paper with few remarks.  An additional dataset has been used for experiment and we report some results in the first subsection. Then we remark on how to obtain more labeled image data to further improve our proposed methods.



\subsection{More Experimental Results}

\begin{figure*}[t]
  \centering
  \begin{tabular}{cc}
 \includegraphics[width=.45\columnwidth]{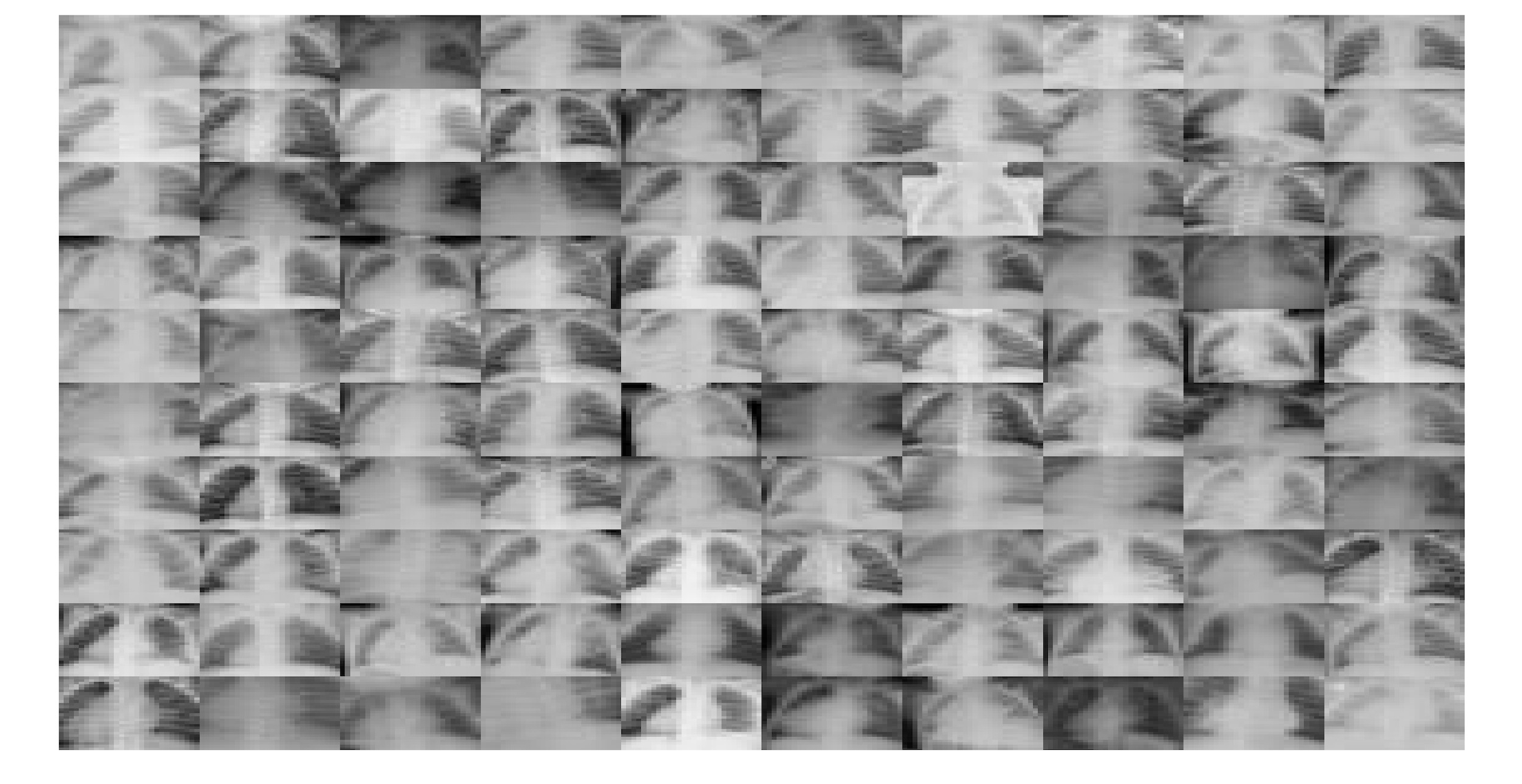} &
        \includegraphics[width=0.45\columnwidth]{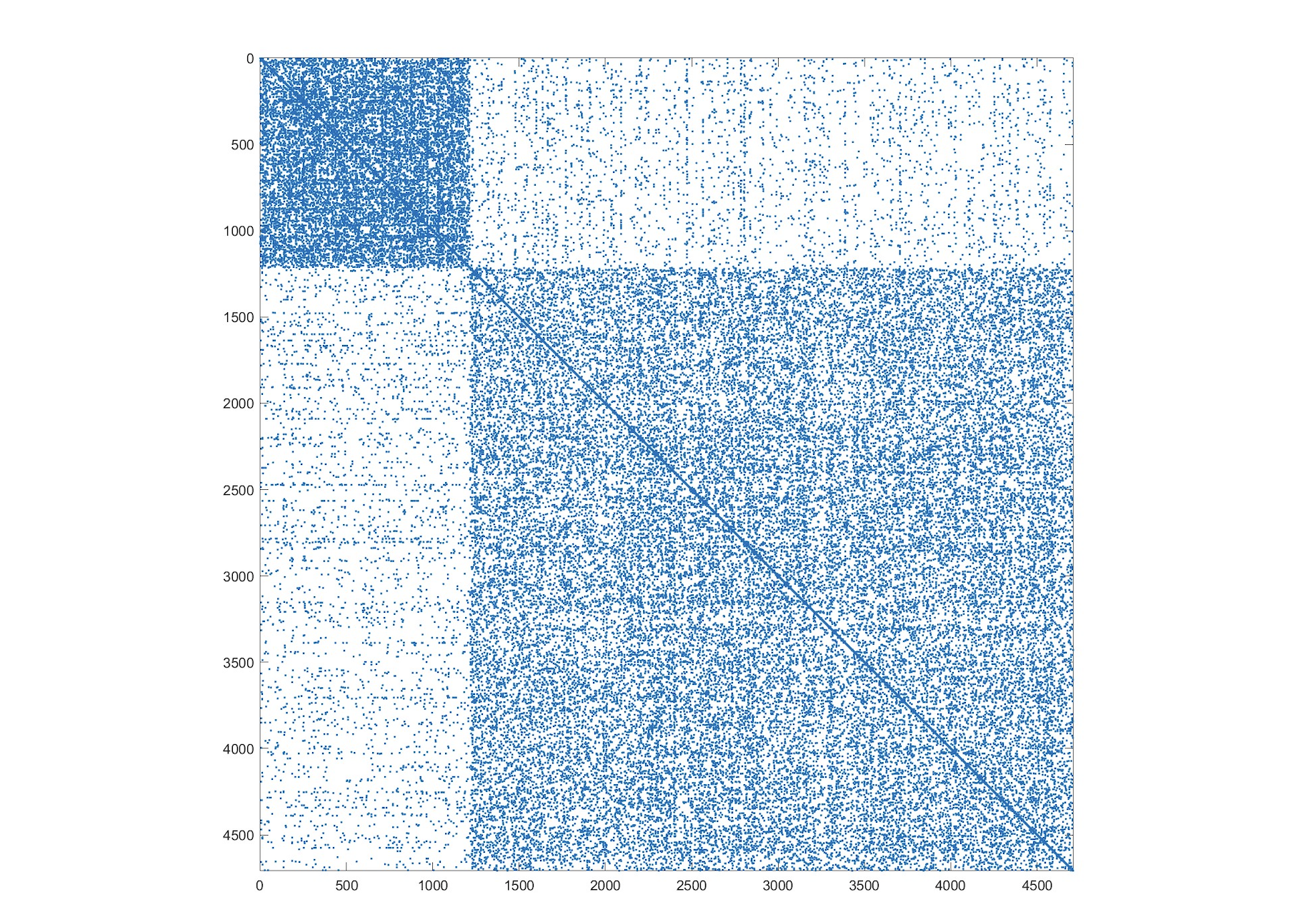}
        \end{tabular}
        \caption{A Few Sample Images (left) and an Adjacency Matrix $A$ (right) associated with the Pneumonia Image Data \label{PneumoniaA}}
\end{figure*}

\begin{figure*}[h]	
	\centering
       
\begin{tabular}{cc}
     \includegraphics[width=.45\columnwidth]{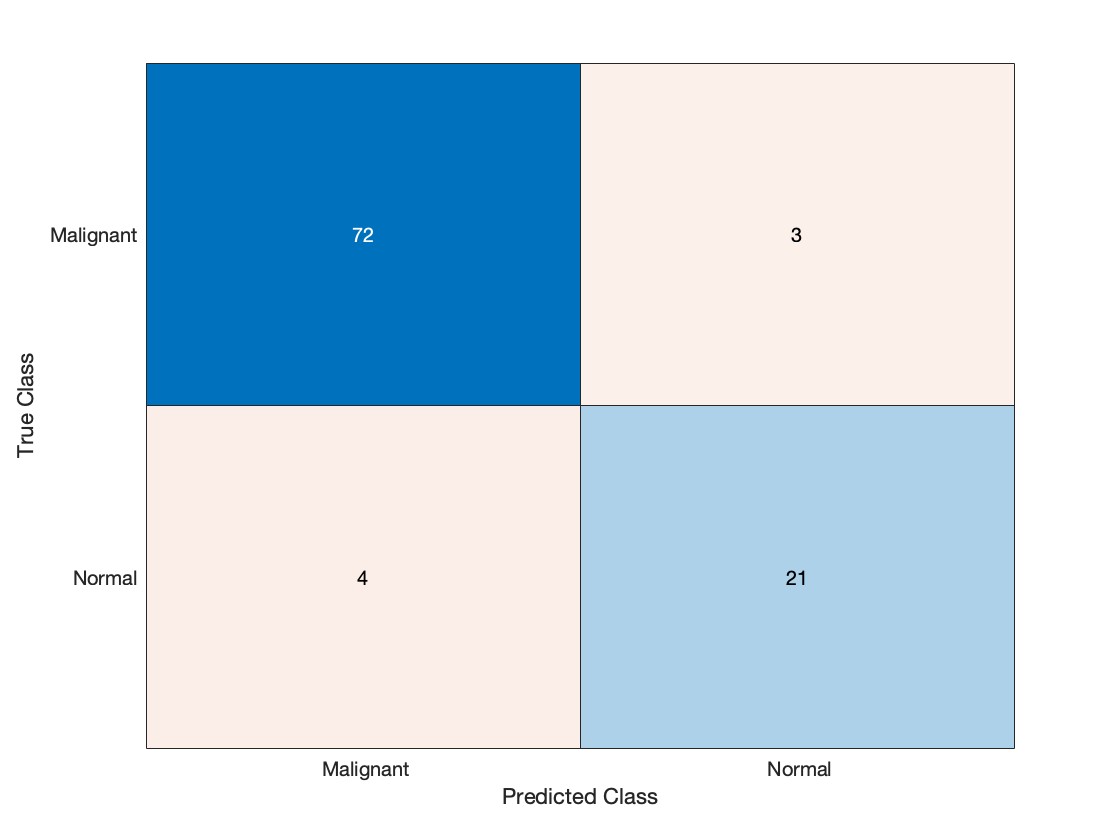}   
     & \includegraphics[width=.45\columnwidth]{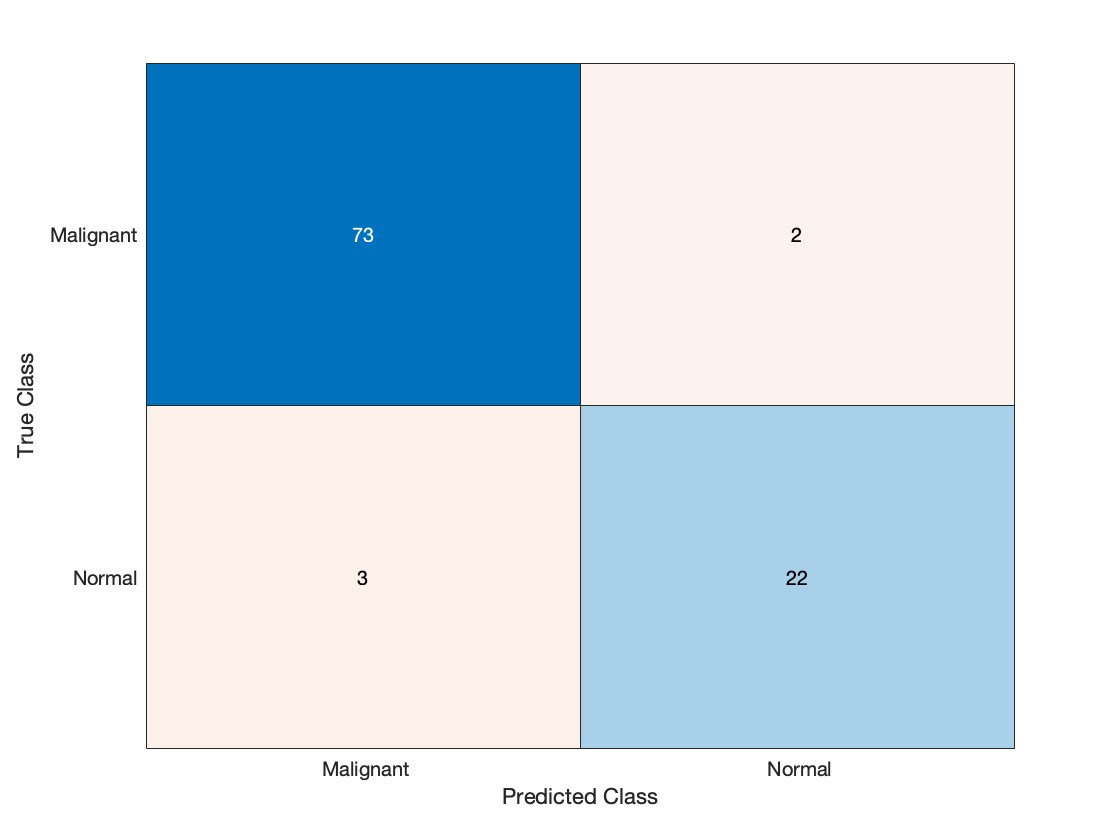} 
\end{tabular} 
  \vspace{-3mm}
	\caption{Confusion matrix of LCE (left) and LSC (right) on Pneumonia dataset. The values are rounded to integers. \label{CmatrixPneu}}
\end{figure*}

In addition to the ``IQ-OTH/NCCD" lung cancer dataset, we used the Pneumonia patient images from \cite{Y21}, which is a collection of MNIST-like images for more experiments. The dataset consists 4708 images (we only use the training part of this dataset), where 1214 are normal and 3494 are malignant. Therefore it is a binary classification problem. A small samples of images are shown in
Figure~\ref{PneumoniaA}. 
We again use the modified exponential distance function to generate 
an adjacency matrix, which has very nice block structure as shown in Figure~\ref{PneumoniaA}. Therefore no LN method is needed for this dataset.

We randomly choose 100 images (25 normal, 75 malignant) out of the 4708 images as testing samples. For each testing image, we apply LCE or LSC on the adjacency matrix which is built by using modified exponential distance between the testing image and the remaining images with known labels. The size of output cluster are set to be 8 for LCE (or 10 for LSC), after excluding the seeded testing image, the cluster is of size 7 for LCE (or 9 for LSC).

The labels of these images in the output cluster are used to determine the label of the testing image. If more than half of the labels are normal, we conclude the corresponding testing image as normal. Otherwise, we conclude the corresponding testing image as malignant. 

Figure~\ref{CmatrixPneu} gives the confusion matrix over 100 independent repetitions of the experiment. Note that the entries in the confusion matrix are rounded to be integers. In Table~\ref{tab3}, 
we show the average precision, recall, and F1-score for each category over 100 independent repetitions of the experiment. The overall accuracy ($\pm$ standard deviation) for LCE and LSC on Pneumonia dataset are $93.04\%\pm 2.45\%$ and $94.06\%\pm 2.17\%$ respectively, while the performances of other benchmark approaches on this dataset are auto-sklearn $85.5\%$, AutoKeras $87.8\%$, ResNet50 $88.4\%$, GoogleAutoMLVision $94.6\%$ (cf. \cite{Y21}). Therefore, our proposed local clustering methods can achieve a favorable or comparable results with much less computational cost.


\subsection{More Artificially Generated Labeled Images }


One direction to improve the performance of the proposed LCE and LSC methods is to generate more artificial images, i.e., increase the number of labeled data, hence any given query image will be more likely to find a faithful cluster and hence make a reliable identification.  
We propose two approaches to do so. 

\begin{table}[t]
\centering
\caption{Precision/Recall/F1-score of Image Identification by the two methods on Pneumonia Dataset \label{tab3}}
\begin{tabular}{ccc}
\toprule
          & Malignant   & Normal  \\
    \midrule
	LCE &   95.99 / 94.68 / 95.33 & 84.67 / 88.12 / 86.38   \\
 LSC & 96.05 / 97.33 / 96.69   & 89.68 / 86.16 / 87.88 \\
	\bottomrule

\end{tabular}
\vspace{-5mm}
\end{table} 

\begin{figure*}[h]
  \centering
       \includegraphics[width=.48\columnwidth, height=.2\columnwidth]{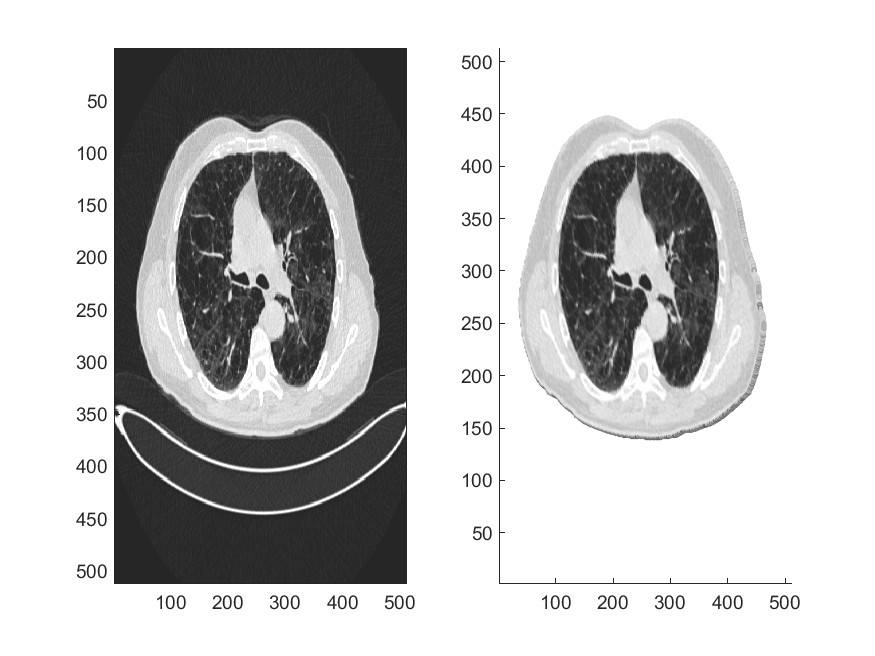}
        \includegraphics[width=.48\columnwidth, height=.2\columnwidth]{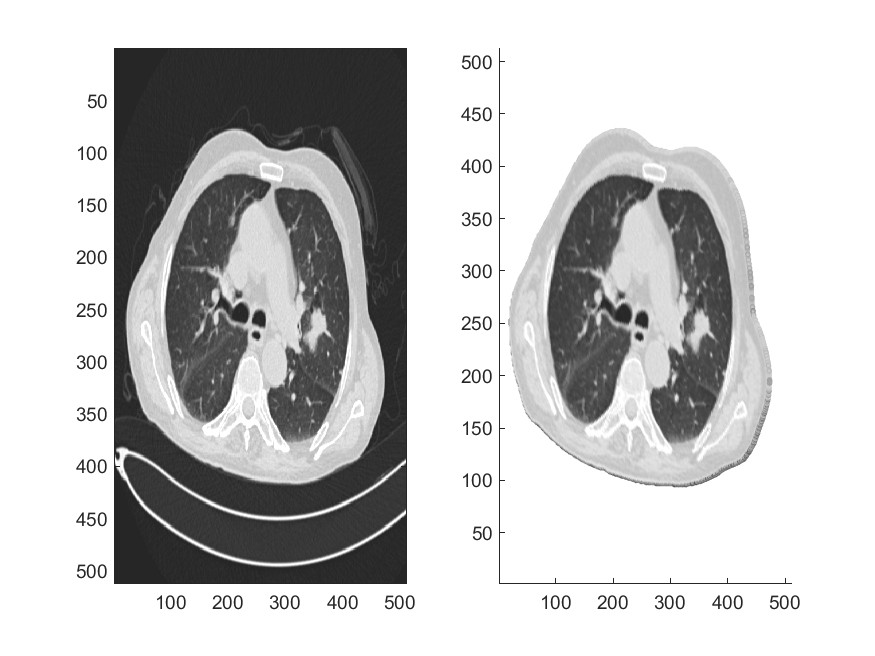}
        \vspace{-1mm}
        \caption{The chest of a test image (leftmost), the domain $V$ of interest (second from left), the chest of a malignant image  (third from left) and the domain $W$ of interest (rightmost) }
        \label{LungCancerDeformation}
    \end{figure*}

\begin{figure*}[h]
  \centering
       \includegraphics[width=.48\columnwidth, height=.2\columnwidth]{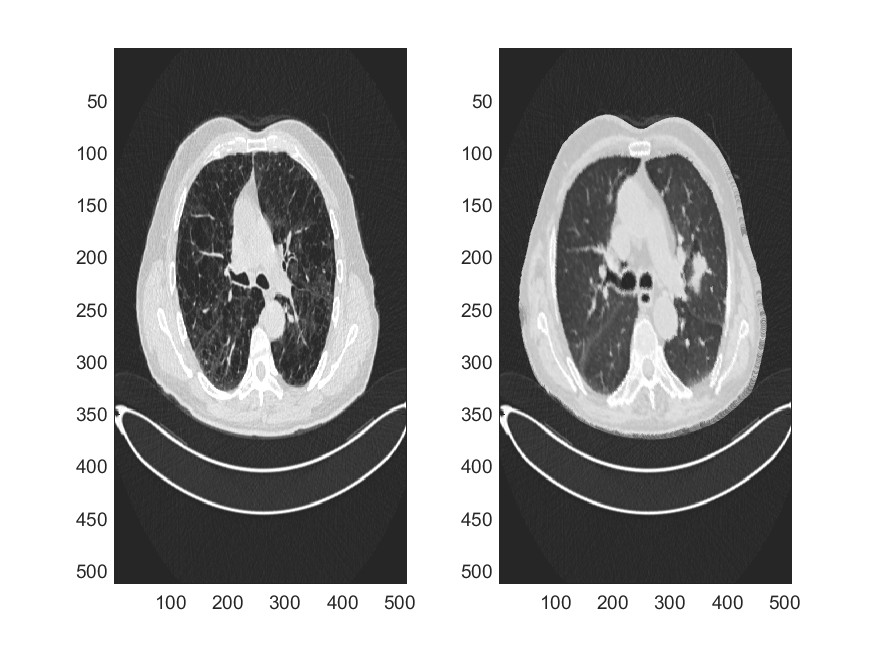}
        \includegraphics[width=.48\columnwidth, height=.2\columnwidth]{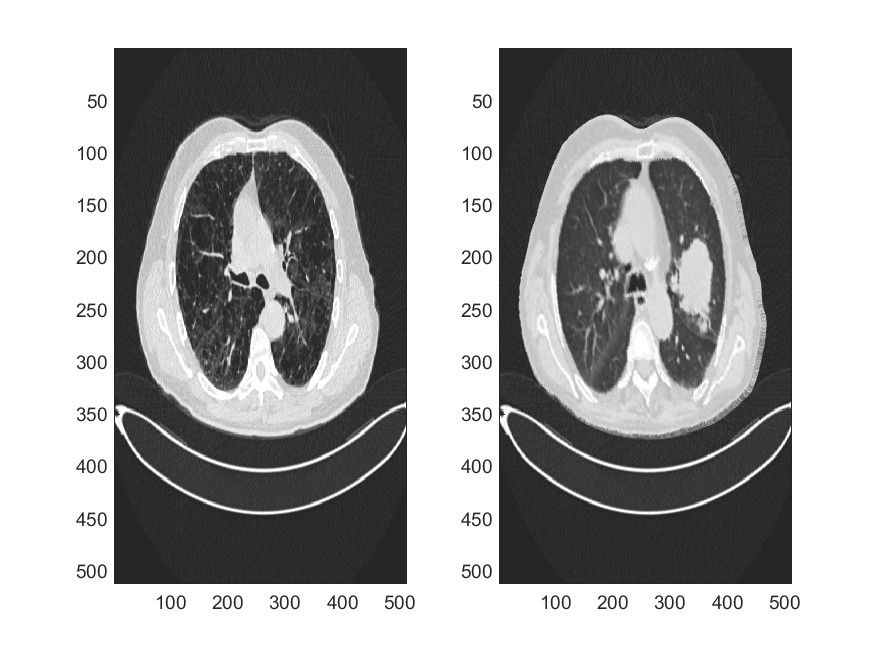}
        \vspace{-1mm}
        \caption{The  chest of the test image (leftmost), the chest with malignant lung (second from left) after the GBC deformation from the most right image of Figure~\ref{LungCancerDeformation},  
        and another pair of the test image and the deformed image using another malignant lung (rightmost). \label{LungCancerDeformation2}}
    \end{figure*}

\begin{figure*}[h]
  \centering
       \includegraphics[width=.48\columnwidth, height=.2\columnwidth]{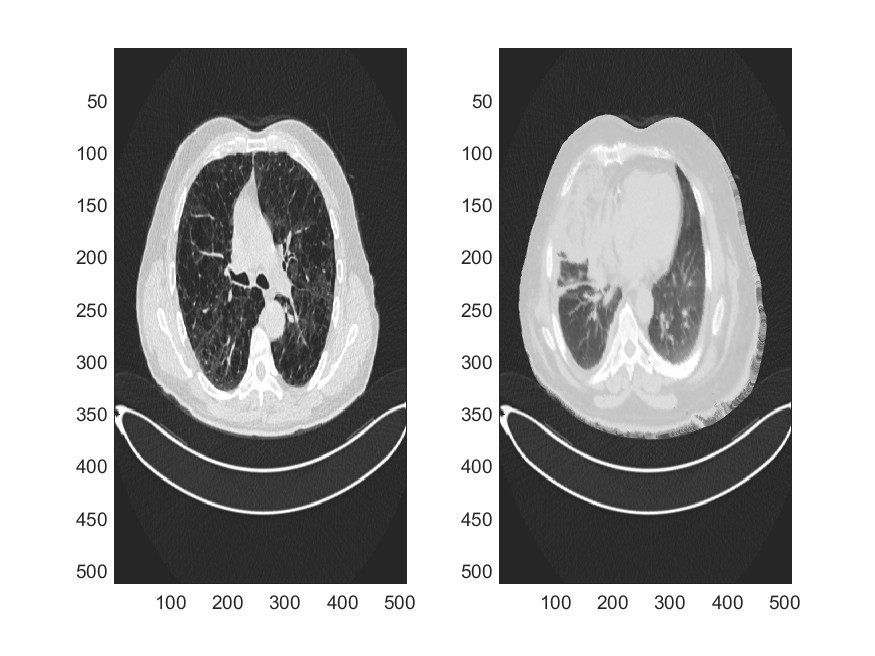}
        \includegraphics[width=.48\columnwidth, height=.2\columnwidth]{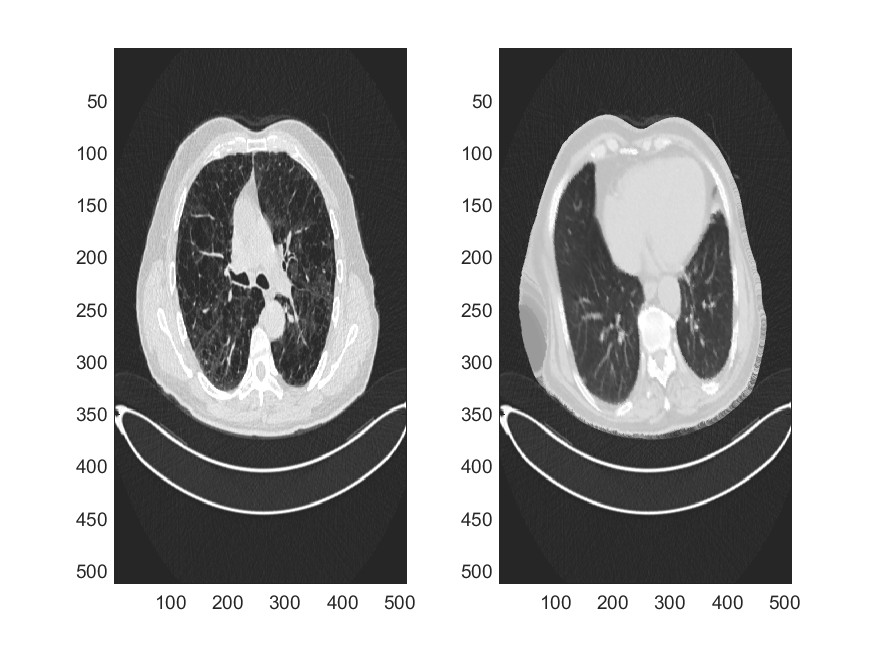}
        \vspace{-1mm}
        \caption{More examples of the chest images with malignant lungs deformed to the chest of the test image. \label{LungCancerDeformation3}}
    \end{figure*}

One approach for generating more labeled data is based on the numerical solution to the Monge-Amp\'ere equation (cf. \cite{LL24}) for the optimal transport problem. That is,  we deform the portion over the chest of each image in the malignant class to the portion of the chest in each image in the normal class. There are a few computational difficulties to overcome such as identifying the centers of the domains of the chests. We shall leave the details to a future publication. 
Another approach is to use the method of harmonic generalized
barycentric coordinates (GBC) to do 
the image deformation. See \cite{DHL22} and \cite{H23}  
for some examples of image deformation. Let us illustrate this 
idea by Figures \ref{LungCancerDeformation}, \ref{LungCancerDeformation2}, and \ref{LungCancerDeformation3}.    

The idea is, we first isolate the chest of each medical image as shown in Figure~\ref{LungCancerDeformation} from the most left one 
to the second left one and similarly from the third left to the most right one.  Let $V$ be the domain
of interest in the second from left in Figure~\ref{LungCancerDeformation}, and $W$ be the domain on the 
right of Figure~\ref{LungCancerDeformation}.  We deform $W$ to $V$ by using the so-called harmonic GBC method as discussed in 
(cf. \cite{DHL22} and \cite{H23}). This results in the 
image on the second from the left of  Figure~\ref{LungCancerDeformation2}. We can see that the resulting image is a new chest image with a malignant lung. Similarly, we can use the method to obtain many more examples of malignant lung cases in different chest images via deformation, as shown in Figure~\ref{LungCancerDeformation2}
and Figure~\ref{LungCancerDeformation3}.   
We can do such deformation between all pairs of malignant, benign, and normal images. In this way, we will obtain more than ten thousand labeled images. 

If a test image has a similar chest to one of the images in the normal class but has a malignant lung resembling that in one of malignant cases, our local clustering methods will be able to identify it as a malignant case. Even if a testing image has a chest different in silhouette from all existing labeled images, we can deform all the chests from the images in the malignant class to the chest of the testing image. Then we use our approach to find if the testing image should be labeled a malignant image or not.  

The goal of the idea above is to deform, via the GBC deformation or Optimal transportation, more than 500 chest images with malignant lungs to the chest of the
query image. Our method will determine whether the lung area in the testing image resembles any of the malignant cases, much like the process of doctors making diagnoses based on their experiences, i.e.
the memory of thousands of images of malignant lungs, the deformation of these malignant lungs into the  chest of the query image, and the comparison of the testing image to make a diagnosis. On the other hand, we will have more than 10,000 labeled images by deforming all malignant lungs and begnin lungs to each of normal lungs as artificially labeled images for training. One of the main research problems is to identify some landmark points along the boundary of the chest so that the deformation can be done accurately. 
We leave the generation of landmarks of these images over the entire dataset to future work.







\end{document}